\documentstyle[named]{iwcs}

\title{An Implemented Formalism for Computing Linguistic
  Presuppositions and Existential Commitments}

\author{Daniel Marcu and Graeme Hirst\\
Department of Computer Science \\ University of Toronto \\ Toronto,
Ontario  \\ Canada \hspace{1em} M5S 1A4 \\{\tt \{marcu,gh\}@cs.toronto.edu}
}

\newcommand{\comment}[1]{}

\newcommand{\uentails}{\models^u}

\newcommand{\dentails}{\models^d}

\newtheorem{definition}{Definition}[section]

\newcommand{\name}[1]{\label{#1}}
\newcommand{\sent}[1]{(\ref{#1})}

\newcounter{stce}[part]
\newcounter{next}[part]
\newcounter{nextone}[part]
\newcounter{lastone}[part]
\def\stce
    {  \refstepcounter{stce}
       \stepcounter{next}
       \stepcounter{nextone}
       \stepcounter{lastone}
       \begin{itemize}
       \item[{\bf (\thestce )}~]%
    }

\def\endstce
    {\end{itemize}}

\def\example#1{\stce#1\endstce}

\def\next{(\thenext)}

\newcounter{fstce}[section]
\renewcommand{\thefstce}{\roman{fstce}}
\def\fstce
    {
       \refstepcounter{fstce}
       \begin{itemize}
       \item[(\thefstce)~]
    }
\def\endfstce
    {\end{itemize}}

\begin{document}

\maketitle

\begin{abstract}
We rely on the strength of linguistic and philosophical perspectives
in constructing a framework that offers a unified explanation for
presuppositions and existential commitment.  We use a rich ontology
and a set of methodological principles that embed the essence of
Meinong's philosophy and Grice's conversational principles into a
stratified logic, under an unrestricted interpretation of the
quantifiers. The result is a logical formalism that yields a tractable
computational method that uniformly calculates all the presuppositions
of a given utterance, including the existential ones.
\end{abstract}

\begin{multicols}{2}

\section{Introduction}

It is common knowledge that a rational agent is inclined to presuppose
the existence of definite references that occur in utterances. Hearing
or uttering the examples below, a rational agent presupposes that the
cheese, children, and car physically exist.
\example{The cheese I bought yesterday is very bad.}
\example{I really don't know what to do with my children anymore.}
\example{Sorry I couldn't make it; my car broke on my way.}
However, day-to-day English provides an impressive number of cases
when existential presuppositions are not inferred, or when they are
defeated by some commonsense knowledge \mbox{(see~\cite{hirst91}}
for a comprehensive study).  One can explicitly speak of
nonexistence~\sent{nonexistence}; events and actions that do not
occur~\sent{not occur}; existence at other times~\sent{other times};
or fictional and imaginary objects~\sent{fictions}.
\example{No one got an A+ in this course. \name{nonexistence}}
\example{John's party is cancelled. \name{not occur}}
\example{G\"odel was a brilliant mathematician. \name{other times}}
\example{Sherlock Holmes is smarter than any other detective.
\name{fictions}}
Note that the simple dichotomy found in most approaches to
presupposition between existent and nonexistent objects is not enough
for a full account of natural language expressiveness.

The study of presuppositions is primarily a study of commitment ---
commitment to the existence of presupposed definite referents or to
the truth of factive complements.  The reduction of presupposition to
entailment is inadequate because presuppositions are {\em implied},
not {\em specified}; they are not part of the truth conditions of
natural language sentences, and they can be cancelled in negative
environments.   Trying to explain the whole phenomenon and to provide
solutions for the projection problem, linguists have often omitted any
explanation for the existential commitment of definite references or
their explanation has been a superficial one.  Similarly, philosophers
who have studied existence and nonexistence have been more concerned
with providing formal tools for manipulation of nonexistent objects
than tools to capture our commonsense commitment.  This puts us in a
difficult position. From a linguistic perspective, the literature
provides a good set of theories able to more or less explain the
commitment to the presupposed truth of factives and the like but not
the existential commitment of definite references.  From a
philosophical perspective, we have quite a few theories which deal
with existence and nonexistence, but they too offer no explanation for
existential commitment.

Our aim here is to provide a formalism that has the
strength of both perspectives. We achieve this using the following:
\begin{itemize}
\item a set of methodological principles that unify
Meinong's~\shortcite{meinong04} philosophy with
Grice's~\shortcite{grice75} conversational principles;
\item  a rich ontology in the style of Hirst~\shortcite{hirst91},
which  provides  the possibility of having consistent models that
contain objects belonging to different ontological spaces;
\item an extension of stratified logic~\cite{marcu94}
where  the quantifiers are read under
Lejewski's~\shortcite{lejewski54}   ``unrestricted interpretation'',
which provides us the formal tool for expressing the above layers.
\end{itemize}

The implementation relies on an extension of the Beth semantic
tableaux to stratified logic. The code is written in Common Lisp and
makes extensive use of the nondeterministic facilities of the Screamer
system~\cite{siskind93b,siskind93a}.

We first review the philosophical approaches in studying existence
and nonexistence and the linguistic approaches in studying
presuppositions, emphasizing their (in)ability to deal with
presuppositions and nonexistence respectively. We give a brief
introduction to stratified logic, its implementation, and explain the
methodological principles of our approach.  In section 4 we show how
this approach is able not only to deal with nonexistence but also able to
explain the existential commitment of definite reference. The rest of
the paper is dedicated to a comparison with Parsons's and Hobbs's
work.

\section{What philosophers and linguists have to say}
\subsection{ Nonexistence and commitment in philosophy and logic}

Early works of Frege~\shortcite{frege92} and
Russell~\shortcite{russell05} tackled a very small subset of what
today is labelled with the name ``presupposition'': the
presuppositions introduced by definite references and proper names.
Hirst~\shortcite{hirst91} shows that classical logic, which embeds
Quine's metaphysical view~\cite[p.150]{quine49} that ``everything
exists'', is not able to deal adequately with nonexistent objects. For
example, if one knows that dragons do not exist --- $(\forall x)(\neg
dragon(x))$ --- it is impossible to distinguish between {\em My dragon
has blue eyes} and {\em My dragon does not have blue eyes} because
both translations in first-order logic are false: $(\exists
x)(dragon(x) \wedge my(x) \wedge [\neg]has\_blue\_eyes(x))$.
Therefore, first-order logic is doomed to fail in any attempt to
reflect the presuppositions of definite references.

Several approaches to nonexistence rely on Meinong's {\em mental act}
philosophy~\cite{meinong04}.  For example,
Parsons~\shortcite{parsons80} avoids Russell's paraphrase of the
definite description by using the predicate $E!$. For Parsons, $(\iota x)
\Phi$ refers to the unique object that satisfies $\Phi$ if there is
such an object. Otherwise, it does not refer to anything at all.  For
a sentence such as ``The man in the doorway is clever'',
Parsons~\shortcite[p.114]{parsons80} argues that the translation
$(\iota x)((man(x) \wedge in\_the\_doorway(x)) \wedge clever(x))$ is not
adequate because it does not reflect our commitment to the man's
existence. Therefore, he proposes that the translation should be
$(\iota x)((E!(x) \wedge man(x) \wedge in\_the\_doorway(x)) \wedge
clever(x))$ where $E!$ is the existential predicate. But the problem
with this is that it embeds the existential commitment in the logical
translation --- not as something that is ``implied'' or presupposed,
but as something ``said'' or specified.  This is not the case with
linguistic presuppositions. Thus, the first translation is too weak
--- unable to capture the commitment, and the second one is too strong
--- the commitment becomes part of the translation and leaves no room
for cancellation of presupposition.

Outside Meinong's world, we find other approaches that focus on the
appropriate reading of the quantifiers.
Lejewski~\shortcite{lejewski54} and Hintikka~\shortcite{hintikka59}
both propose an ``unrestricted interpretation'' of the quantifiers,
which makes no commitment to the existence of the objects over which
they range.  Under this interpretation, existence can be predicated
(Lejewski), or explicitly captured as $(\exists x)(x = a)$ (Hintikka).
The latter solution is nothing but a translation into logic of Quine's
slogan, ``to exist is to be the value of a variable''.  In these
universes, we are free now to talk about Pegasus and dragons but we
cannot explain our commitment to the existence of the definite
referents.

An interesting approach towards explaining the conditions in which
existential presuppositions are generated is built by
Atlas~\shortcite{atlas88} around the notions of ``aboutness'' and
``noun-phrase topicality''.  Instead of allowing all the noun-phrases
in a sentence to exhibit presupposition generation capabilities, only
the topical ones enjoy this property. Atlas gives no hint of how this
theory could be extended to deal with factives or verbs of judging, and
defining the notions of aboutness and topicality for them is not
trivial. Even if we did manage to do this, such presuppositions can
never be cancelled. Either they are generated or they are not. This
leads us to believe that sentences such as {\em John didn't stop
beating the rug because he never started} cannot be captured in this
manner.

\label{hobbs1}

Hobbs~\shortcite{hobbs85} uses the ``unrestricted interpretation'' of
the quantifiers introduced by Lejewski~\shortcite{lejewski54}. Hence,
in Hobbs's framework, the set of things we can talk about (including,
therefore, nonexistent things) and the set of things we quantify over
are equal. The existential commitment is captured by a set of
``transparency axioms''. For example, the sentence {\em Ross worships
Zeus} is represented as:

\[
        worship^\prime(E,Ross,Zeus) \wedge Exist(E) \label{ross worships zeus}
\]
The first conjunct says that $E$ is the event of worshipping Zeus by
Ross,  and the second says that $E$ exists in the real world. Hobbs
assigns a {\em transparency} property to the predicates. For
$worship^\prime$, this property entails the existence of its second
argument in the physical world:
\[
\begin{array}{l}
\forall E \forall x \forall y ((worship^\prime(E,x,y) \wedge Exist(E)) \\
\hspace*{10mm} \rightarrow Exist(x)) \label{worship' transparency}
\end{array}
\]

Apparently, the commitment to Ross's existence is solved.
$Worship^\prime$ is transparent in its second argument but not in its
third; so we may infer that Ross is existent, but draw no conclusions
about Zeus.  The problem is that the transparency axioms are
associated with the predicates and not with the objects, so that there
is no criterion to choose an appropriate translation for a sentence
like {\em The King of Buganda worships Zeus} because the translation
should be transparent if we know nothing about Buganda and opaque
otherwise.

\subsection{Theories of linguistic presupposition  and their relation to
(non)existence}

The vast majority of the linguistic approaches are more concerned with
``how presuppositions are inherited'' than with ``what presuppositions
are''. Presuppositions are defined in terms of {\em plugs, holes,} and
{\em filters}~\cite{karttunen73}, consistency~\cite{gazdar79},
uncontroversiality~\cite{soames82}, or hypothetical and secondary
contexts~\cite{kay92}, but nothing is said about the logical framework
into which they may be expressed. An exception is Mercer's
approach~\shortcite{mercerphd}.  He abandons the projection method in
favour of rules of inference in default logic. Our main objection is
to Mercer's use of natural disjunction as an exclusive disjunction,
and the reduction of natural implication to logical equivalence.
Mercer~\shortcite{mercer93} argued that this is a consequence of the
way he intended his ``proof by cases'', in which ``the cases are taken
from a conjunctive statement, where the conjuncts are the disjuncts in
a classical proof''. He assumes that this non-standard notion is the
one that must be used in nonmonotonic reasoning.  But this
non-traditional analysis and the reduction of natural implication to
logical equivalence are not representable within the logic itself.
Hence, this method is also a procedural one.

A different perspective is given by Sandt~\shortcite{sandt92} and
Zeevat~\shortcite{zeevat92} for whom presuppositions are understood as
anaphoric expressions that have internal structure and semantic
content. Because they have more semantic content than other anaphors,
presuppositions are able to create an antecedent in the case that the
discourse does not provide one. Van der Sandt provides a computational
method for presupposition resolution in an enhanced discourse
representation theory, while Zeevat gives a declarative account for it
using update semantics, but neither of the methods is able to
accommodate the cancellation of presupposition that is determined by
information added later to the discourse. A simple ontology consisting
only of existent and nonexistent objects is inadequate for dealing
with fictions or objects that have unactualized existence. Therefore,
sentences such as {\em Sherlock Holmes is smarter than any other
detective} or {\em The strike was averted} cannot be represented in
their theories.

\section{ Reasoning in stratified logic}

Stratified logic~\cite{marcu94} reflects a different understanding of
default reasoning phenomena from that found in the classic
literature~\cite{reiter80}. Instead of treating the notion of
defeasibility on consistency and justification-based grounds, we
conjecture that defeasible inferences are ``weaker'' than classical
entailments. For the purpose of this paper, it is enough to consider
only a subset of stratified logic.

In first-order stratified logic, a stratified interpretation ${\cal
SL}$ consists of an universe of objects $U$ and a function mapping $F$
as in first-order logic, but the relation set is partitioned according
to the strength (undefeasible and defeasible relations) and polarity
(positive and negative relations).  Thus, the set of relations $R$
will be given by the union $R^u \cup \overline{R^u} \cup R^d \cup
\overline{R^d}$ where $R^u$ stands for positive undefeasible
relations, $\overline{R^u}$ for negative undefeasible relations, $R^d$
for positive defeasible relations, and $\overline{R^d}$ for negative
defeasible relations. Positive atomic formulas and negative (negated)
atomic formulas are labelled as defeasible (e.g. $p^d(t_1,\ldots,t_n)$)
or undefeasible (e.g. $\neg p^u(t_1,\ldots,t_n)$) and compound
formulas are obtained from positive and negative atomic formulas using
classical logical connectors.  For example, one would formalize that
uttering that {\em John does not regret that Mary came to the party}
presupposes that {\em Mary came to the party} as
\begin{equation}
\begin{array}{l}
\neg regret^u(John,come(Mary,party)) \rightarrow \\
\hspace*{10mm} come^d(Mary,party) \label{regret1}
\end{array}
\end{equation}
because {\em Mary came to the party} is defeasible: {\em John does not
regret that Mary came to the party because she did not come}.

At the semantic level, we extend the notion of satisfiability to the
two levels we have introduced; hence, we will have {\em
u-satisfiability}, $\uentails$, and {\em d-satisfiability}, $\dentails$.

\begin{definition} \label{xentailsfol}
Assume $\sigma$ is an  ${\cal SL}$ valuation such that $t_i^\sigma =
d_i \in {\cal D}$ and assume that ${\cal SL}$  maps $n$-ary predicates
$p$ to relations $R \subset {\cal D} \times \ldots \times {\cal D}$.
For any atomic formula $p^x(t_1,\ldots,t_n)$, and any stratified
valuation $\sigma$, where $x \in \{u,d\}$ and
$t_i$ are terms, the x-satisfiability relations are
defined as follows:

\begin{itemize}
\item $\sigma \uentails p^u(t_1,\ldots,t_n)$ iff \hspace*{3cm}
\linebreak \hspace*{2cm} $\langle d_1, \ldots, d_n \rangle \in R^u$
\item $\sigma \uentails p^d(t_1,\ldots,t_n)$ iff \hspace*{3cm}
\linebreak  \hspace*{2cm} $\langle d_1, \ldots, d_n \rangle \in R^u
\cup \overline{R^u} \cup R^d$
\item $\sigma \dentails p^u(t_1,\ldots,t_n)$ iff \hspace*{3cm}
\linebreak  \hspace*{2cm} $\langle d_1, \ldots, d_n \rangle \in R^d$
\item $\sigma \dentails p^d(t_1,\ldots,t_n)$ iff \hspace*{3cm}
\linebreak \hspace*{2cm} $\langle d_1, \ldots, d_n \rangle \in R^d $
\end{itemize}

For any negation of an atomic formula $\neg p^x(t_1,\ldots,t_n)$,
and any stratified valuation $\sigma$, where $x
\in \{u,d\}$ and $t_i$ are terms, the x-satisfiability relations are
defined as follows:

\begin{itemize}
\item $\sigma \uentails \neg p^u(t_1,\ldots,t_n)$ iff \hspace*{3cm}
\linebreak \hspace*{2cm} $\langle d_1, \ldots, d_n \rangle \in
\overline{R^u}$
\item $\sigma \uentails \neg  p^d(t_1,\ldots,t_n)$ iff \hspace*{3cm}
\linebreak \hspace*{2cm} $\langle d_1, \ldots, d_n \rangle \in R^u
\cup \overline{R^u} \cup \overline{R^d}$
\item $\sigma \dentails \neg  p^u(t_1,\ldots,t_n)$ iff \hspace*{3cm}
\linebreak \hspace*{2cm} $\langle d_1, \ldots, d_n \rangle \in
\overline{R^d}$
\item $\sigma \dentails \neg  p^d(t_1,\ldots,t_n)$ iff \hspace*{3cm}
\linebreak \hspace*{2cm} $\langle d_1, \ldots, d_n \rangle \in
\overline{R^d}$
\end{itemize}
\end{definition}

The {\em x-satisfiability} relation for compound formulas is defined
in the usual way. One can see that this definition of satisfiability
has two major advantages: on one hand, the $\uentails$ relation
provides a high degree of liberty in satisfying sets of formulas that
contain positive and negative information of different strengths; on
the other hand the $\dentails$ relation is able to signal when such a
contradiction occurs. For example, in accordance with the above
definition, the theory $\{ \neg p^u(t_1,\ldots,t_n),
p^d(t_1,\ldots,t_n) \}$ is {\em u-satisfiable} but is not {\em
d-satisfiable}. That means defeasible and undefeasible information are
allowed to co-exist because the satisfiability relations are able to
handle them appropriately.

Stratified logic uses an extension of semantic tableaux that is both
sound and complete to compute the models associated with a given
theory. On a set of model schemata, we define a partially ordered
relation ($\leq$) that yields the {\em most optimistic} schemata for the
theory, i.e., those that contain more information and whose
information is as defeasible as possible. For example, a translation
in stratified logic of the classical example involving Tweety
(represented by the constant $T$) will yield three model schemata.
Schema $m_1$ does not cancel the fact that Tweety flies as schema
$m_2$ does.  Moreover, $m_1$ contains more information than $m_3$.
Therefore, $m_1$ is the most optimistic model schema.
\[
\left\{
\begin{array}{l}
bird^u(T) \\
(\forall x)(bird^u(x) \rightarrow flies^d(x)) \\
(\forall x)(penguin^u(x) \rightarrow bird^u(x)) \\
(\forall x)(penguin^u(x) \rightarrow \neg flies^u(x)) \\
\end{array}
\right.
\]

\noindent
\begin{tabular}{|l|l|l|} \hline \hline
{\em Schema\#} & {\em Indefeasible} &  {\em Defeasible} \\ \hline
$m_1$ & $bird^u(T)$      &        \\
      & $\neg penguin^u(T)$  &        \\
      &          &  $flies^d(T)$ \\ \hline
$m_2$ & $bird^u(T)$      &        \\
      & $\neg penguin^u(T)$  &        \\
      & $\neg flies^u(T)$   &  $flies^d(T)$ \\ \hline
$m_3$ & $bird^u(T) $      &        \\
      & $\neg flies^u(T)$ & $flies^d(T)$ \\ \hline
\end{tabular}
\vspace{3mm}

\noindent Model schema $m_1$ corresponds to an ${\cal SL}$ structure defined
over an universe that contains only one object, $T$, and no
function symbols. The relations defined on the universe are $R^u = \{
bird(T) \}$, $\overline{R^u} = \{ penguin(T) \}$ and $R^d =
\{ flies(T) \}$. For the sake of compactness and clarity we
represent stratified models as unions of relations partitioned according
to their strength:
\[
\begin{array}{l}
m_1 = \{ bird^u(T),\neg  penguin^u(T) \} \hspace*{1mm} \cup  \{
flies^d(T) \} \\
\end{array}
 \]

The stratified semantic tableaux and the model-ordering relations have
been fully implemented in Common Lisp using the Screamer macro package
that provides nondeterministic facilities
{}~\cite{siskind93b,siskind93a}. Our program takes as input a
logical representation of the background knowledge and of an
utterance, computes the model schemata for the theory, and returns the set of
most optimistic schemata and the presuppositions associated with a given
utterance in the case that they exist. Computing the model schemata for a
stratified theory can be done within the same complexity bounds as in
first-order logic. The algorithm for determining the most optimistic
schemata is $O(n^2)$.

\section{ Presuppositions as defeasible information}

\subsection{Methodological principles for our approach}

The approach to nonexistent objects and presuppositions that we are going
to present is constructed on the basis of a modified  set of Meinongian
principles about nonexistence. They are embedded in a stratified logic
framework in which quantifiers are taken under Lejewski's unrestricted
interpretation. The ontology is enhanced with the eight types of
existence listed by Hirst~\shortcite{hirst91}, though in this paper, we
will deal only with physical existence, represented as $E!$,
unactualized existence, represented as $U\!E!$, existence
outside the world but with causal interaction with that world,
$EOW!$, and existence in fiction, $F!$.

Following Rapaport's style~\shortcite{rapaport85a}, we propose a set
of methodological principles based on Meinong~\shortcite{meinong04}
that are meant to capture the ability of an intelligent agent to deal
with existence and nonexistence rather from a conversational
perspective than from a rational one.

\begin{description}
        \item[MC1.] Every uttered sentence is ``directed'' towards an
``object'', because every uttered sentence can be seen as a
materialization of a mental act.

        \item[MC2.] All uttered sentences  exist (technically, ``have being'').
However, this does not imply the existence of their referents,  which
are ``ausserseiend'' (beyond being and non-being).

        \item[MC3.] It is not self-contradictory to deny, nor tautologous to
affirm, the existence of a referent.

        \item[MC4.] Every referent and every uttered sentence  has properties.

        \item[MC5.] The principles MC2 and MC4 are not inconsistent.

        Corollary: Even referents of an uttered sentence that do not
exist have properties.

        \item[MC6.] (a) Every set of properties (Sosein) corresponds to the
utterance of a sentence. \\
                    (b) Every object of thought can be uttered.

        \item[MC7.] Some referents of an utterance are incomplete (undetermined
with
respect to some properties).
\end{description}

In accordance with Grice~\shortcite{grice75}, we need two additional
principles:
\begin{description}
\item[GC1.] The speaker is committed to the truth of the sentences he utters.

\item[GC2.] Using and deriving  presuppositions requires, from both speaker and
listener, a sort of ``optimism''.
\end{description}
Principle GC1 is formalized by the translation of the uttered
sentences into classical logic formulas in which quantifiers are read
under their unrestricted interpretation. Principle GC2 is formalized
by the rules containing defeasible information that exist in the
knowledge base of the speaker and the hearer, and the notion of
optimism in the model-ordering relation. For example, a factive
negation weakly implies the truth of its complement (see
formula~\ref{regret1} above).  Note that a non-optimistic
interpretation of utterances will never be able to account for any of
the pragmatic inferences, because they are not explicitly uttered.

\subsection{Formalizing presuppositions}

We assume that our inference process relies not only on {\em core
knowledge} as in ``all men are mortal'' or ``birds fly'', but also on
{\em knowledge of language use} as in ``a factive negation weakly implies
the truth of its complement'' as shown in formula~\ref{regret1}.

That definite references imply the existence of their referents
constitutes another instance of defeasible inference (see
examples~\sent{nonexistence}---\sent{fictions}).  We can capture this
either by adding a new formula \[ (\forall x)({\mbox {\sf
definite\_reference(x)}} \rightarrow E!^d(x)) \] to our knowledge
base, and by embedding syntactic terms into the logical form, as Hobbs
did~\shortcite{hobbs85}, or by representing this defeasible commitment
explicitly in the translation of each utterance containing a definite
reference or proper noun.  Both approaches exhibit the same semantic
behavior, and due to the model-ordering relation they explain our
commitment to a referent's existence (in the case that we do not know
otherwise). Because ${\mbox {\sf definite\_reference(x)}}$ is
syntactic information, we depict it using a different font, but the
reader should understand that ${\mbox {\sf x}}$ is bound by the same
quantifier as $x$ is, and that  ${\mbox {\sf definite\_reference(x)}}$
is used as a metalogical symbol that triggers pragmatic inferences.

As a last step, we abandon the Fregean reading of the quantifiers
and we adopt Lejewski's unrestricted
interpretation~\shortcite{lejewski54}. This means that $\exists$ and
$\forall$ do not mix quantification with ontological commitment:
$(\exists x)object(x)$ does not entail the physical existence of $x$,
so the things we can talk about equals the things we can quantify
over.  This yields the following:

{\bf Definition:} {\em Presuppositions} are  defeasible information
that is derived from knowledge of language use and that is included in
the most optimistic models of a theory described in terms of stratified
logic under an unrestricted interpretation of the quantifiers.

\subsection{What the approach can do with existent and nonexistent objects}

Assume that someone utters the sentence {\em The king of Buganda is
(not) bald.} If we know nothing about Buganda and its king, the
complete theory of this utterance and the available knowledge in
stratified logic is this:
 \[
\left\{
\begin{array}{l}
(\exists x)(king\_of\_buganda^u(x) \wedge \\
\hspace*{5mm} {\mbox {\sf definite\_reference(x)}} \wedge (\neg)
bald^u(x)) \\
(\forall x)({\mbox {\sf definite\_reference(x)}} \rightarrow E!^d(x))
\\
\end{array} \right.
\]
This theory has one optimistic model that reflects  one's commitment to
the king's existence. The king's existence has the status of {\em defeasible
information}; it is derived using knowledge of language use and is a
presupposition of the utterance.
\[
\begin{array}{l}
m = \{king\_of\_buganda^u(\xi_0), (\neg)bald^u(\xi_0)\} \\
\hspace*{10mm} \cup \hspace{1mm} \{E!^d(\xi_0)\}
\end{array}
\]

Knowledge about the political system of France can inhibit
the inference regarding the existence of its king in a sentence such as
{\em The king of France is (not) bald}. Assume that  we know
there is no king of France $(\neg E!^u)$.  A complete formalization follows:
\[
\left\{
\begin{array}{l}
(\exists x)(king\_of\_france^u(x) \wedge {\mbox {\sf
definite\_reference(x)}} \\
\hspace*{5mm}  \wedge (\neg) bald^u(x)) \\
(\forall x)({\mbox {\sf definite\_reference(x)}} \rightarrow E!^d(x))
\\
(\forall x)( king\_of\_france^u(x) \rightarrow \neg E!^u(x)) \\
\end{array} \right.
\]
For this theory, we obtain only one model schema:

\vspace{5mm}
\noindent
\begin{tabular}{|l|l|l|} \hline \hline
{\em Schema \#} & {\em Indefeasible} &  {\em Defeasible} \\ \hline
$m_1$ & $king\_of\_france^u(\xi_o)$      &        \\
      & $(\neg) bald^u(\xi_0)$  &        \\
      & $\neg E!^u(\xi_0)$ & $E!^d(\xi_0)$ \\ \hline
\end{tabular}
\vspace{3mm}

\noindent One can notice that the existential presupposition is now
cancelled by some background knowledge. The only way one can satisfy
the initial theory is if she has a stratified structure where $\neg
E!^u(\xi_0)$.  Thus, the theory yields one model
\[
\begin{array}{l}
m = \{king\_of\_france^u(\xi_0), (\neg)bald^u(\xi_0), \\
\hspace*{10mm}  \neg E!^u(\xi_0)\} \hspace{1mm}  \cup  \hspace{1mm} \O^d
\end{array}
\]

Asserting existence
or nonexistence affects {\em defeasible inferences} due to knowledge of
language use and restricts some of the models.
If someone utters {\em The king of Buganda exists} and  we know nothing
about Buganda, the translation
\[
\left\{
\begin{array}{l}
(\exists x)(king\_of\_buganda^u(x) \wedge {\mbox {\sf
definite\_reference(x)}} \\
\hspace*{5mm}  \wedge E!^u(x)) \\
(\forall x)({\mbox {\sf definite\_reference(x)}} \rightarrow E!^d(x))
\end{array}
\right.
\]
gives  one  model:
\[
\begin{array}{l}
m = \{ king\_of\_buganda^u(\xi_0),  E!^u(\xi_0) \} \hspace{1mm}  \cup
\hspace{1mm} \O^d \\
\end{array}
\]
If we {\em know} that the king of Buganda does not exist, or in other
words we evaluate the above sentence against a knowledge base that
contains
\[ (\forall x)(king\_of\_buganda^u(x) \rightarrow \neg E!^u(x)), \]
there is no model for this theory, so the utterance is interpreted as
false.  It is noteworthy that the inconsistency appears due to
specific knowledge about the king's physical existence and not because
of a quantification convention as in classical first-order logic. On
the other hand, the negation, {\em The king of Buganda does not
exist}, is consistent with the knowledge base and provides this model:
\[ m = \{king\_of\_buganda^u(\xi_0), \neg E!^u(\xi_0)\} \hspace{1mm}  \cup
\hspace{1mm} \O^d \]

So far, we have emphasized the way presuppositions of definite
references can be handled in this framework. However, the proposed
method is general in the sense that it captures the other
presuppositional environments as well.  Moreover, the cancellation can
occur at any moment in discourse. Consider for example the
utterance {\em John does not regret that Mary came to the party}. Its
formalization in stratified logic follows:
\[
\left\{
\begin{array}{l}
(\neg regret^u(john,come(mary,party)) \wedge \\
\hspace*{3mm} {\mbox {\sf definite\_reference(john)}} \wedge \\
\hspace*{3mm} {\mbox {\sf definite\_reference(mary)}} \wedge \\
\hspace*{3mm} {\mbox {\sf definite\_reference(party)}} \\
(\forall x)({\mbox {\sf definite\_reference(x)}} \rightarrow E!^d(x))
\\
(\forall x,y,z)(\neg regret^u(x, come(y,z)) \\
\hspace*{3mm} \rightarrow come^d(y,z)) \\
\end{array} \right.
\]
The optimistic model computed by our program is this:
\[
\begin{array}{l}
m = \{ \neg regret^u(john,come(mary,party)) \} \hspace{1mm} \cup \\
\hspace*{10mm} \{ E!^d(john), E!^d(mary), E!^d(party), \\
\hspace*{12mm} come^d(mary,party) \}
\end{array}
\]
This model reflects our intuitions that {\em Mary came to the
party} and all definite references exist.

If one utters now {\em Of course he doesn't. Mary did not come to the
party}, the new model computed by our program will reflect the fact that a
presupposition has been cancelled, even though this cancellation
occurred later in the discourse. Thus, the new optimistic model will
be this:
\[
\begin{array}{l}
m = \{ \neg regret^u(john,come(mary,party)), \\
\hspace*{11mm}  \neg come^u(mary,party) \} \hspace{1mm} \cup  \\
\hspace*{10mm}   \{ E!^d(john), E!^d(mary), E!^d(party) \}
\end{array}
\]

Our approach correctly handles references to unactualized objects such
as averted strikes or the paper that we had wanted to submit to AAAI-94. The
utterance {\em The strike was averted} can be formalized thus:
\[
\left\{
\begin{array}{l}
(\exists x)(strike^u(x) \wedge {\mbox {\sf
definite\_reference(x)}} \\
\hspace*{5mm}  \wedge averted^u(x)) \\
(\forall x)({\mbox {\sf definite\_reference(x)}} \rightarrow E!^d(x))
\\
(\forall x)(averted^u(x) \rightarrow U\!E!^u(x)) \\
(\forall x)(U\!E!^u(x) \rightarrow \neg E!^u(x)) \\
\end{array} \right.
\]
This gives one optimistic model:
\[
\begin{array}{l}
m = \{strike^u(\xi_0), averted^u(\xi_0), \neg
E!^u(\xi_0), \\
\hspace*{10mm} U\!E!^u(\xi_0) \} \hspace*{1mm} \cup \hspace{1mm}  \O^d  \\
\end{array}
\]

\section{A comparison with Parsons's and Hobbs's work}

\subsection{On Parsons's evidence for his theory of nonexistence}

Parsons argues that is impossible to distinguish between the shape of
the logical form of two sentences like these, in which one subject is
fictional and the other is real:
\begin{quote}
        a. Sherlock Holmes is more famous than any other detective.

        b. Pel\'{e} is more famous than any other soccer player.
\end{quote}
In our approach, similar syntactic translations give different
semantic models when interpreted against different knowledge
bases. A complete theory for the first sentence is this:
\[
\left\{
\begin{array}{l}
(\exists x)(sherlock\_holmes^u(x) \wedge {\mbox {\sf
definite\_reference(x)}}   \\
\hspace*{3mm} \wedge (\forall y)((detective^u(y) \wedge (x \neq y))
\rightarrow \\
\hspace*{5mm} more\_famous^u(x,y)))   \\
(\forall x)({\mbox {\sf definite\_reference(x)}} \rightarrow E!^d(x)) \\
(\forall x)( sherlock\_holmes^u(x) \rightarrow F!^u(x)) \\
(\forall x)( F!^u(x) \rightarrow \neg E!^u(x))
\end{array}
\right.
\]
This theory gives only one model:
\[
\begin{array}{l}
m = \{sherlock\_holmes^u(\xi_0), \neg E!^u(\xi_0),
F!^u(\xi_0 ), \\
\hspace*{10mm} detective^u(y), more\_famous^u(\xi_0,y)\} \cup \O^d
\end{array}
\]
This corresponds to an object $\xi_{0}$ that does not exist in the
real world but exists as a fiction, has the property of being Sherlock
Holmes, and for any other object $y$, real or fictional that has the
property of being a detective, the object $\xi_{0}$ is more famous
than object $y$.  Of course, in this model, it is impossible to commit
ourselves to Holmes's physical existence, but is possible to talk
about him.

The theory for the second sentence is this:
\[
\left\{
\begin{array}{l}
(\exists x)(pele^u(x) \wedge {\mbox {\sf definite\_reference(x)}}  \wedge \\
\hspace*{3mm} (\forall y)((soccer\_player^u(y) \wedge
(x \neq y)) \rightarrow \\ \hspace*{5mm} more\_famous^u(x,y))) \\
(\forall x)({\mbox {\sf definite\_reference(x)}} \rightarrow E!^d(x)) \\
\end{array}
\right.
\]
This theory exhibits one optimistic model:
\[
\begin{array} {l}
m =  \{pele^u(\xi_0), soccer\_player^u(y), \\
\hspace*{10mm} more\_famous^u(\xi_0,y)\} \hspace*{1mm} \cup \hspace*{1mm}
\{ E^d!(\xi_0) \}   \\
\end{array}
\]
Model $m$ states that the object $\xi_{0}$, being Pel\'{e}, exists in a
defeasible sense and this is the existential presupposition of the
initial utterance.

As seen, it is needless to mention the existence of specific objects
in the knowledge base.  The model-ordering relation rejects anyhow
models that are not optimistic. In this way, the commitment to Pel\'{e}'s
existence is preserved, and appears as a presupposition of the
utterance. Parsons's theory provides different logical forms for the
above sentences, but fails to avoid the commitment to nonexistent
objects.


\subsection{A comparison with Hobbs's work}

We have mentioned that Hobbs's transparency pertains to relations and
not to objects. In our approach, a sentence such as {\em Ross worships
Zeus} can be satisfied by a set of semantic models that correspond to
each possible combination of the existence and non-existence of Ross
and Zeus.
\[
\left\{
\begin{array}{l}
(\exists x)(\exists y)(ross^u(x) \wedge zeus^u(y) \wedge worship^u(x,y)
 \\
\hspace*{3mm} \wedge {\mbox {\sf definite\_reference(x)}} \wedge {\mbox {\sf
definite\_reference(y)}}) \\
(\forall x)({\mbox {\sf definite\_reference(x)}} \rightarrow E!^d(x))
\end{array} \right.
\]
Among them, only one is minimal: the one that explains the commitment
to both  Ross's  and Zeus's existence.
\[
\begin{array}{l}
m = \{ross^u(\xi_0), zeus^u(\xi_1), worship^u(\xi_0,\xi_1) \} \\
\hspace*{10mm} \cup \hspace*{1mm} \{ E!^d(\xi_0), E!^d(\xi_1) \}
\end{array}
\]
But let us  assume we know that there is no  entity in the real world that
enjoys the property of being Zeus, but rather one who exists outside the real
world as a god $(EOW!^u)$.
\[
\left\{
\begin{array}{l}
(\exists x)(\exists y)(ross^u(x) \wedge zeus^u(y) \wedge worship^u(x,y)
 \\
\hspace*{3mm} \wedge {\mbox {\sf definite\_reference(x)}} \wedge {\mbox {\sf
definite\_reference(y)}}) \\
(\forall x)({\mbox {\sf definite\_reference(x)}} \rightarrow E!^d(x)) \\
(\forall x)(zeus^u(x) \rightarrow EOW!^u(x)) \\
(\forall x)(EOW!^u(x) \rightarrow \neg E!^u(x)) \\
\end{array} \right.
\]
This theory is no longer satisfiable by a model in which Zeus exists
as a physical entity. However, the optimistic model explains our
commitment to Ross's existence.
\[
\begin{array}{l}
m = \{ross^u(\xi_0), zeus^u(\xi_1), \neg E!^u(\xi_1), EOW!^u(\xi_1), \\
\hspace*{10mm} worship^u(\xi_0,\xi_1) \} \hspace*{1mm}  \cup
\hspace*{1mm}  \{ E!^d(\xi_0) \}
\end{array}
\]

\section{Conclusion}

Joining Meinong's philosophy of nonexistence with Grice's
conversational principles provides a very strong motivation for a
uniform treatment of linguistic presuppositions. Lejewski's
unrestricted interpretation of the quantifiers, Hirst's ontology, and
the notion of reasoning with stratified tableaux and
model-ordering in stratified logic provide the formal
tools to implement the principles. This amounts to a model-theoretic
definition for presuppositions that is able to offer a uniform
treatment for linguistic presuppositions and an explanation for the
existential commitment. A computationally tractable method can be
derived from the formalism. Its implementation in Common Lisp finds
the natural language presuppositions, including the existential ones,
and correctly reflects their cancellation.

\vspace{6mm}

\noindent{\large\bf Acknowledgements}

\vspace{3mm}

\noindent This research was supported in part by a grant from the Natural
Sciences and Engineering Research Council of Canada.


\end{multicols}

\end{document}